\documentclass[11pt]{article}
\usepackage[a4paper]{geometry}
\usepackage{clic2017}
\usepackage{times}
\usepackage{url}
\usepackage{latexsym}
\usepackage{graphicx}
\usepackage{tabularx}
\usepackage{booktabs}
\usepackage{todonotes}
\usepackage{verbatim}
\usepackage{subcaption}
\usepackage{float}

\title{Personality Profiling in Italian Social Media: a Novel MBTI-AnnotatedCorpus}

\title{Personal-ITY: A Novel YouTube-based Corpus  \\for Personality Prediction in Italian}

\author{Elisa Bassignana \\
  Dipartimento di Informatica\\
  University of Turin \\
  {\small {\tt elisa.bassignana@edu.unito.it}} \\\And
  Malvina Nissim \\
  CLCG \\ University of Groningen \\
  {\small {\tt m.nissim@rug.nl}} \\\And
  Viviana Patti \\
  Dipartimento di Informatica\\
  University of Turin \\
  {\small {\tt viviana.patti@unito.it}} \\}

\date{}

\begin{document}

\maketitle
\begin{abstract}
We present a novel corpus for personality prediction in Italian, containing a larger number of authors and a different genre compared to previously available resources. The corpus is built exploiting  Distant Supervision, assigning \emph{Myers-Briggs Type Indicator} (\emph{MBTI}) labels to YouTube comments, and can lend itself to a variety of experiments. We report on preliminary experiments on Personal-ITY, which can serve as a baseline for future  work, showing that some types are easier to predict than others, and discussing the perks of cross-dataset prediction.
\end{abstract}

{\let\thefootnote\relax\footnotetext{Copyright \textcopyright\ 2019 for this paper by its authors. Use permitted under Creative Commons License Attribution 4.0 International (CC BY 4.0).}} 

\section{Introduction}

When faced with the same situation, different humans behave differently. This is, of course, due to different backgrounds, education paths, and life experiences, but according to psychologists there is  another important aspect: personality \cite{snyder1983influence,parks2009personality}.

Human Personality is a psychological construct aimed at explaining the wide variety of human behaviours in terms of a few, stable and measurable individual characteristics \cite{vinciarelli2014survey}.

Such characteristics are formalised in
\textit{Trait Models}, and there are currently two of these models that are widely adopted:
\emph{Big Five} \cite{John1999TheBF} and \emph{Myers-Briggs Type Indicator} (\emph{MBTI}) \cite{myers1995gifts}.
The first examines five dimensions ({\scshape Openness to experience}, {\scshape Conscientiousness}, {\scshape Extroversion}, {\scshape Agreeableness} and {\scshape Neuroticism}) and for each of them assigns a score in a range.
The second one, instead, considers 16 fixed personality types, coming from the combination of the opposite poles of 4 main dimensions ({\scshape Extravert-Introvert}, {\scshape iNtuitive-Sensing}, {\scshape Feeling-Thinking}, {\scshape Perceiving-Judging}). Examples of full personality types are therefore four letter labels such as {\scshape ENTJ} or {\scshape ISFP}.

The tests used to detect prevalence of traits include human judgements regarding semantic similarity and relations between adjectives that people use to describe themselves and others. This is because language is believed to be a prime carrier of personality traits \cite{schwartz2013personality}. This aspect, together with the progressive increase of available user-generated data on social media, has prompted the task of \emph{Personality Detection}, i.e., the automatic prediction of personality from written texts
\cite{youyou2015computer,automatprof,litvinova,whelan2006profiling}. 

Personality detection can be useful in predicting life outcomes such as substance use, political attitudes and physical health. Other fields of application are marketing, politics and psychological and social assessment.

As a contribution to personality detection in Italian, we present Personal-ITY, a new corpus of YouTube comments annotated with MBTI personality traits, and some preliminary experiments to highlight its characteristics and test its potential. The corpus is made available to the community\footnote{\url{https://github.com/elisabassignana/Personal-ITY}}.

\section{Related Work}

There exist a few datasets annotated for personality traits. For the shared tasks organised within the \emph{Workshop on Computational Personality Recognition} 
\cite{celli2013workshop}, two datasets annotated with the \textit{Big Five} traits have been released in 2013 (Essays \cite{essays} and myPersonality\footnote{\url{http://mypersonality.org}}) and two in 2014 (YouTube Personality Dataset \cite{biel2013youtube} and Mobile Phones interactions \cite{mobileDataset}).

For the 2015 PAN Author Profiling Shared Task \cite{Pardo2015OverviewOT}, personality was added to gender and age in the profiling task, with tweets in English, Spanish, Italian and Dutch. These are also annotated according to the \textit{Big Five} model.

Still in the Big~Five landscape, \newcite{schwartz2013personality} collected a dataset of  FaceBook comments (700~millions~words) written by 136.000 users who shared their status updates. Interesting correlations were observed between word usage and personality traits.

If looking at data labelled with the MBTI traits, we find a corpus of 1.2M English tweets annotated with personality and gender \cite{Personality_Traits_on_Twitter}, and the multilingual 
{\scshape TwiSty} \cite{Verhoeven2016TwiStyAM}.
The latter is a corpus of data collected from Twitter annotated with MBTI personality labels and gender for six languages (Dutch, German, French, Italian, Portuguese and Spanish) and a total of  18,168 authors. We are interested in the Italian portion of {\scshape TwiSty}.

Table \ref{tab:corpus} contains an overview of the available Italian corpora labelled with personality traits. We include our own, which is described in Section~\ref{sec:corpus}.

\begin{table}
\centering
\begin{tabularx}{0.45\textwidth}{Xlrr}
\toprule
Corpus & Model & \# user & Avg. \\
\toprule
PAN2015 & Big Five & 38 & 1258 \\
{\scshape TwiSty} & MBTI & 490 & 21.343 \\
Personal-ITY & MBTI & 1048 & 10.585 \\
\bottomrule
\end{tabularx}
\caption{Summary of Italian corpora  with personality labels. Avg.: average tokens per user.}
\label{tab:corpus}
\end{table}

Regarding detection approaches, 
\newcite{mairasse} tested the usefulness of different sets of textual features making use of mostly SVMs.

At the PAN~2015~challenge (see above) a variety of algorithms were tested (such as Random Forests, decision trees, logistic regression for classification, and also various regression models), but overall most successful participants used SVMs. Regarding features, 
participants approached the task with combinations of style-based and content-based features, as well as their combination in \emph{n}-gram models \cite{Pardo2015OverviewOT}.

Experiments on {\scshape TwiSty} were performed by the corpus creators themselves using a LinearSVM with  word (1-2) and character (3-4) \emph{n}-grams. Their results (reported in Table~\ref{expTS} for the Italian portion of the dataset) are obtained through 10-fold cross-validation; the model is compared to a weighted random baseline (WRB) and a majority baseline (MAJ).
\begin{table}[h]
\centering
\begin{tabularx}{0.45\textwidth}{l|XXX}
\toprule
Trait & WRB & MAJ & f-score \\
\toprule
EI & 65.54 & 77.88 & 77.78\\
NS & 75.60 & 85.78 & 79.21\\
FT & 50.31 & 53.95 & 52.13\\
PJ & 50.19 & 53.05 & 47.01\\
\midrule
Avg & 60.41 & 67.67 & 64.06 \\
\toprule
\end{tabularx}
\caption{{\scshape TwiSty} scores from the original paper. Note that all results are reported as \textit{micro}-average F-score.}
\label{expTS}
\end{table}

\section{Personal-ITY}
\label{sec:corpus}
First, we explain two major choices that we made in creating Personal-ITY, namely the source of the data and the trait model. Second, we describe in detail the procedure we followed to construct the corpus. Lastly, we provide a description of the resulting dataset.

\paragraph{Data} YouTube is the source of data for our corpus. The decision is grounded on the fact that compared to the more commonly collected tweets, YouTube comments can be longer, so that 
users are freer to express themselves without limitations. Additionally, there is a substantial amount of available data on the YouTube platform, which is easy to access thanks to the free YouTube APIs. 

\paragraph{Trait Model} Our model of choice is the MBTI. 
The first benefit of this decision is that this model is easy to use in association with a Distant Supervision approach (just checking if a message contains one of the 16 personality types; see Section~\ref{sec:ds}).
Another benefit is related to the existence of {\scshape TwiSty}. Since both {\scshape TwiSty} and Personal-ITY implement the MBTI model, analyses and experiments over personality detection can be carried out also in a cross-domain setting. 

\subsection*{Ethics Statement}

Personality profiling must be carefully evaluated from an ethical point of view. In particular, often, personality detection involves ethical dilemmas regarding appropriate utilization and interpretations of the prediction outcomes \cite{ethicalconsiderations}. Concerns have been raised regarding the inappropriate use of these tests with respect to invasion of privacy, cultural bias and confidentiality \cite{mehta2019recent}.

The data included in the Personal-ITY dataset were publicly available on the YouTube platform at the time of the collection. As we will explain in detail in this Section, the information collected are comments published under public videos on the YouTube platform by authors themselves.
For a major protection of user identities, in the released corpus only the YouTube usernames of the authors are mentioned which are not unique identifiers.
The YouTube IDs of the corresponding channels, which are the real identifiers in the platform, allowing to trace the identity of the authors, are not released.
Note also that the corpus was created for academic purposes and is not intended to be used for commercial deployment or applications.

\subsection{Corpus Creation}
\label{sec:ds}

The fact that users often self-disclose information about themselves on social media makes it possible to adopt \textit{Distant Supervision} (DS) for the acquisition of training data. DS is a semi-supervised method that has been abundantly and successfully used in affective computing and profiling to assign silver labels to data on the basis of indicative proxies \cite{go2009twitter,pool2016distant,emmery-etal-2017-simple}.

Users left comments to some videos on the MBTI theory in which they were stating their own personality type
(e.g. \emph{Sono ENTJ...chi altro?} [en: "I'm ENTJ...anyone else?"]). We exploited such comments to create Personal-ITY with the following procedure.




First, we searched for as many Italian YouTube videos about MBTI as possible, ending up with a selection of ten
 with a conspicuous number of comments as the ones above\footnote{Links to the 10 YouTube videos:\\\tiny{\url{https://www.youtube.com/watch?v=VCo9RlDRpz0}}
\\\tiny{\url{https://www.youtube.com/watch?v=N4kC8iqUNyk}}
\\\tiny{\url{https://www.youtube.com/watch?v=Z8S8PgW8t2U}}
\\\tiny{\url{https://www.youtube.com/watch?v=wHZOG8k7nSw}}
\\\tiny{\url{https://www.youtube.com/watch?v=lO2z3_DINqs}}
\\\tiny{\url{https://www.youtube.com/watch?v=NaKPl_y1JXg}}
\\\tiny{\url{https://www.youtube.com/watch?v=8l4o4VBXlGY}}
\\\tiny{\url{https://www.youtube.com/watch?v=GK5J6PLj218}}
\\\tiny{\url{https://www.youtube.com/watch?v=9P95dkVLmps}}
\\\tiny{\url{https://www.youtube.com/watch?v=g0ZIFNgUmoE}}
}.


Second, we retrieved all the comments to these videos using an AJAX request, and built a list of authors and their associated  MBTI label. A label was associated to a user if they included an MBTI combination in one of their comments. 
Table~\ref{tab:associationMBTI} shows some examples of such associations.
The association process is an approximation typical of DS approaches. To assess its validity, we manually checked 300 random comments to see whether the mention of an MBTI label was indeed referred to the author's own personality. We found that in 19 cases (6.3\%) our method led to a wrong or unsure classification of the user's personality (e.g. \emph{O tutti gli INTJ del mondo stanno commentando questo video oppure le statistiche sono sbagliate :-)}). We can assume that our dataset might therefore contain about 6-7\% of noisy labels.



\begin{table}
\centering
\begin{tabularx}{0.45\textwidth}{XX}
\toprule
Comment & User - MBTI label\\
\toprule
\emph{Io sono ENFJ!!!} & User1 - ENFJ\\
\midrule
\emph{Ho sempre saputo di essere connessa con Lady Gaga! ISFP!} & User2 - ISFP\\
\bottomrule
\end{tabularx}
\caption{Examples of automatic associations \emph{user} - \emph{MBTI personality type}.}
\label{tab:associationMBTI}
\end{table}


Using the acquired list of authors, we meant to obtain as many comments as possible written by them. The YouTube API, however, does not allow to retrieve all comments by one user on the platform.
In order to get around this problem we relied on video similarities, and tried to expand as much as possible our video collection. Therefore, as a third step, we retrieved the list of channels that feature our initial 10 videos, and then all of the videos within those channels.



Fourth, through a second AJAX request, we downloaded all comments appearing below all videos retrieved through the previous step.

Lastly, we filtered all comments retaining those written by authors included in our original list. This does not obviously cover all comments by a relevant user, but it provided us with additional data per author.

\subsection{Final Corpus Statistics}

For the final dataset, we decided to keep only the authors with a sufficient amount of data. More specifically, we retained only users with at least five comments, each at least five token long.

Personal-ITY includes $96,815$ comments by $1048$ users, each annotated with an MBTI label. The average number of comments per user is $92$ and each message has on average $115$ tokens.

\begin{figure}
\centering
\includegraphics[width=0.48\textwidth]{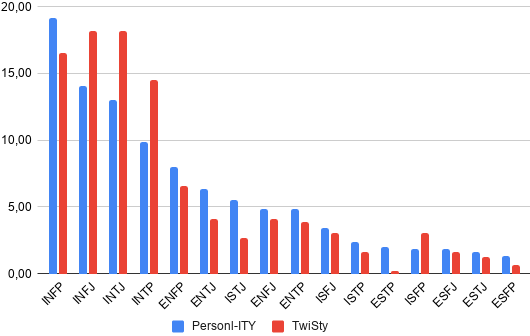}
\caption{Distribution of the 16 personality types in the YouTube corpus and in the Italian section of {\scshape TwiSty}.}
\label{fig:percUsers}
\end{figure}

The amount of the 16 personality types in the corpus is not uniform. Figure~\ref{fig:percUsers} shows such distribution and also compares it with the one in {\scshape TwiSty}. The unbalanced distribution can be due to personality types not being uniformly distributed in the population, and to the fact that different personality types can make different choices about their online presence.
\newcite{PersonalityandonlineofflinechoicesMbtiprofiles} for example, observed that there is a significant correlation between online--offline choices and the MBTI dimension of {\scshape Extravert-Introvert}: 
extroverts are more likely to opt for offline modes of communication, while online communication is presumably easier for introverts. In Figure~\ref{fig:percUsers}, we also see that the four most frequent types are introverts in both datasets.
The conclusion is that, despite the different biases, collecting linguistic data in this way has the advantage that it reflects actual language use and allows large-scale analysis \cite{Personality_Traits_on_Twitter}.

\begin{figure}
\centering
\subfloat[][\emph{Extravert - Introvert}]
{\includegraphics[width=.45\textwidth]{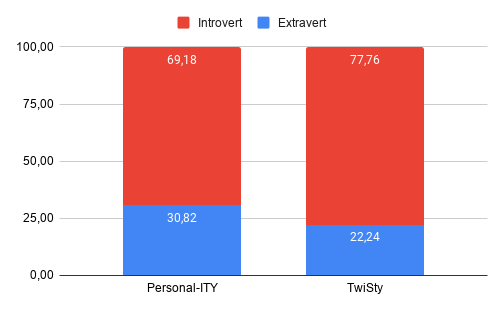}} \\
\subfloat[][\emph{Sensing - Intuitive}]
{\includegraphics[width=.45\textwidth]{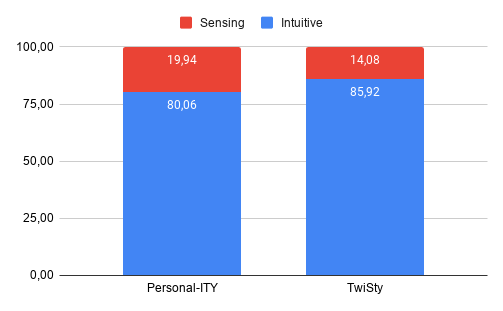}} \\
\subfloat[][\emph{Thinking - Feeling}]
{\includegraphics[width=.45\textwidth]{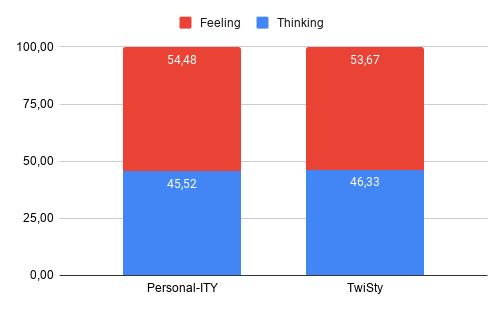}} \\
\subfloat[][\emph{Judging - Perceiving}]
{\includegraphics[width=.45\textwidth]{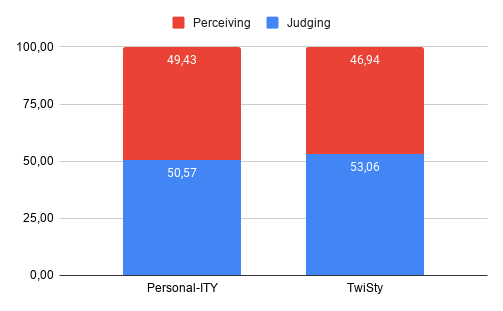}}
\caption{Comparison of the distributions of the four MBTI traits between Personal-ITY and the Italian part of {\scshape TwiSty}.}
\label{fig:countTraits}
\end{figure}

Figure~\ref{fig:countTraits} shows more in detail, trait by trait, the distribution of the opposite poles through the users in Personal-ITY and in {\scshape TwiSty}.
As we might have expected, in line with what is observed in Figure~\ref{fig:percUsers}, the two datasets present very similar trends.
Such similarities between Personal-ITY and {\scshape TwiSty} are these similarities are a further confirmation of the reliability of the data we collected.

\section{Preliminary Experiments}

We ran a series of preliminary experiments on Personali-ITY which can also serve as a baseline for future work on this dataset.
We pre-processed texts by replacing hashtags, urls, usernames and emojis with four corresponding placeholders. We adopted the \verb|sklearn| \cite{scikit-learn} implementation of a linear~SVM (LinearSVM), with standard parameters. We tested three types of features. At the lexical level, we experimented with word (1-2) and character (3-4) \emph{n}-grams, both as raw counts as well as tf-idf weighted. 
Character \emph{n}-grams were tested also with a word-boundary option.
At a more stylistically level, we considered the use of emojis, hashtags, pronouns, punctuation and capitalisation.
Lastly, we also experimented with
embeddings-based representations, by using,
on the one hand, YouTube-specific \cite{embYT} pre-trained models, on the other hand,
more generic embeddings, such as 
the Italian version of GloVe \cite{embGloVe}, which is trained on the Italian Wikipedia\footnote{\url{https://hlt.isti.cnr.it/wordembeddings}}. 
We looked for all the available embeddings of the words written by each author, and used the average as feature. If a word appeared more than once in the string of comments, we considered it multiple times in the final average.

We used 10-fold cross-validation, and assessed the models using macro f-score. Note that the original {\scshape TwiSty} paper uses micro f-score. Thus, for the sake of comparison, we include also micro-F in Table~\ref{expTSnostri} for the MAJ baseline and our lexical n-gram model.
Table~\ref{expYT} shows the results of our experiments with different feature types.\footnote{In Tables~\ref{expYT}--\ref{expTSnostri}, we report the highest scores based on averages of the four traits. Considering the dimensions individually, better results can be obtained by using specific models.} Overall, lexical features (n-grams) perform best. Combining different feature types did not lead to any improvement. Classification was performed with four separate binary classifiers (one per dimension), and with one single classifier predicting four classes, i.e, the whole MBTI labels at once. In the latter case, we observe that the results are quite high considering the increased difficulty of the task.
Table \ref{expTSnostri} reports the scores of our models on {\scshape TwiSty}.
As for Personal-ITY, best results were achieved using lexical features (tf-idf \emph{n}-grams); stylistic features and embeddings are just above the baseline. Our model outperforms the one in \cite{Verhoeven2016TwiStyAM} for all traits (micro-F).

\begin{table}[h]
\centering
\begin{tabularx}{0.45\textwidth}{X|XXXXX}
\toprule
Trait & MAJ & Lex & Sty & Emb & FL \\
\toprule
EI & 40.55 & 51.85 & 40.46 & 40.55 & 51.65 \\
NS & 44.34 & 51.92 & 44.34 & 44.34 & 49.04 \\
FT & 35.01 & 50.67 & 36.27 & 35.01 & 50.86 \\
PJ & 29.49 & 50.53 & 51.04 & 47.06 & 51.03 \\
\midrule
Avg & 37.35 & \textbf{51.24} & 43.03 & 41.74 & 50.65 \\
\bottomrule
\end{tabularx}
\caption{Results of the experiments on Personal-ITY. FL:  prediction of the full MBTI label at once, with a character \emph{n}-gram model.}
\label{expYT}
\end{table}

\begin{table}
\centering
\resizebox{\columnwidth}{!}{
\begin{tabularx}{0.5\textwidth}{X|XX|XXXX}
\toprule
& \multicolumn{2}{c|}{micro~F} & \multicolumn{4}{c}{macro~F}\\\midrule
Trait & MAJ & Lex & MAJ & Lex & Sty & Emb \\
\toprule
EI & 77.75 & \textbf{79.18} & 43.69 & 55.23 & 43.69 & 43.69 \\
NS & 85.92 & \textbf{85.92} & 46.15 & 46.15 & 46.15 & 46.15 \\
FT & 53.67 & \textbf{55.31} & 34.79 & 52.98 & 35.34 & 34.70 \\
PJ & 53.06 & \textbf{54.08} & 34.56 & 53.01 & 35.20 & 34.90 \\
\midrule
Avg & 67.6 & \textbf{68.62} & 39.80 & \textbf{51.84} & 40.09 & 39.86 \\
\bottomrule
\end{tabularx}
}
\caption{Results of our experiments on {\scshape TwiSty}.}
\label{expTSnostri}
\end{table}

To test compatibility of resources and to assess model portability, we also ran cross-domain experiments on Personal-ITY and {\scshape TwiSty}.
In the first setting, we tested the effect of merging the two datasets on the performance of models for personality detection, 
maintaining the 10-fold cross-validation setting and by using the model performing better on average for YouTube and Twitter data (a character \emph{n}-grams model). Table~\ref{expCrossUnited} contains the result of such experiments\footnote{Prediction of the full label at once.}. Scores are almost always lower compared to the in-domain experiments (excepts for NS as regards Twitter scores reported in Table~\ref{expTSnostri}: 46.15 $\rightarrow$ 48.31), but quite increased compared to the majority baseline. 

\begin{table}[H]
\centering
\begin{tabular}{l|c|c}
\toprule
Trait & MAJ & Lex \\
\toprule
EI & 41.64 & 50.57 \\
NS & 44.93 & 48.31 \\
FT & 35.04 & 51.31 \\
PJ & 30.66 & 48.24 \\
\midrule
Avg & 38.07 & 49.61 \\
\bottomrule
\end{tabular}
\caption{Merging Personal-ITY with {\scshape TwiSty}.}
\label{expCrossUnited}
\end{table}

In the second setting, instead, 
we divided both corpora in fixed training and test sets with a proportion of 80/20 and ran the models using lexical features, in order to run a cross-domain experiment. For direct comparison, we run the model in-domain again using this split. Results are shown in Table~\ref{expCrossDomain}.
Cross-domain scores are obtained with the best in-domain model.\footnote{Better results can be obtained with other specific models.}
They drop substantially compared to in-domain, but are always above the baseline.

\begin{table}
\centering
\resizebox{\columnwidth}{!}{
\begin{tabularx}{0.5\textwidth}{X|XXX|XXX}
\toprule
 Train & \multicolumn{3}{c|}{Personal-ITY} & \multicolumn{3}{c}{\scshape TwiSty} \\ \midrule
  Test & \multicolumn{1}{c}{IN} &  \multicolumn{2}{c|}{CROSS} & \multicolumn{1}{c}{IN} & \multicolumn{2}{c}{CROSS} \\
  \cline{2-7}
  
  &  Pers & MAJ  &  {\scshape Twi} & {\scshape Twi}  &  MAJ &  Pers \\
\toprule
EI & 58.94 & 44.94 & 49.33 & 55.66 & 44.59 & 44.59 \\
NS & 52.88 & 47.87 & 47.31 & 47.87 & 45.31 & 45.31\\
FT & 49.20 & 37.58 & 47.09 & 65.26 & 39.13 & 51.04\\
PJ & 54.43 & 32.41 & 32.50 & 56.87 & 36.56 & 38.54 \\
\midrule
Avg & 53.86 & 40.70 & 44.06 & 56.42 & 41.40 & 44.87\\
\bottomrule
\end{tabularx}
}
\caption{Results of the cross-domain experiments. MAJ =  baseline on the cross-domain testset.
}
\label{expCrossDomain}
\end{table}

\section{Conclusions}

The experiments show that there is no single best model for personality prediction, as the feature contribution depends on the dimension considered, and on the dataset.
Lexical features perform best, but they tend to be strictly related to the context in which the model is trained and so to overfit.
 
The inherent difficulty of the task itself is confirmed and deserves further investigations, as assigning a definite personality is an extremely subjective and complex task, even for humans.

Personal-ITY is made available to further investigate the above and other issues related to personality detection in Italian. The corpus can lend itself to a psychological analysis of the linguistic cues for the MBTI personality traits. On this line, it is interesting to investigate the presence of evidences linking linguistic features with psychological theories about the four considered dimensions ({\scshape Extravert-Introvert}, {\scshape iNtuitive-Sensing}, {\scshape Feeling-Thinking}, {\scshape Perceiving-Judging}). First results in this direction are presented in~\cite{bassignanaPeoples}.

\section*{Acknowledgments}

The work of Elisa Bassignana was  partially carried out at the University of Groningen within the framework of the Erasmus+ program 2019/20.

\bibliographystyle{acl}
\bibliography{biblio}

\begin{thebibliography}{}

\bibitem[\protect\citename{Argamon \bgroup et al.\egroup }2009]{automatprof}
Shlomo Argamon, Moshe Koppel, James~W. Pennebaker, and Jonathan Schler.
\newblock 2009.
\newblock Automatically profiling the author of an anonymous text.
\newblock {\em Commun. ACM}, 52(2):119–123, February.

\bibitem[\protect\citename{Bassignana \bgroup et al.\egroup
  }2020]{bassignanaPeoples}
Elisa Bassignana, Malvina Nissim, and Viviana Patti.
\newblock 2020.
\newblock {Personal-ITY: a YouTube Comments Corpus for Personality Profiling in
  Italian Social Media}.
\newblock In Viviana Patti, Malvina Nissim, and Barbara Plank, editors, {\em
  Proceedings of the Third Workshop on Computational Modeling of People's
  Opinions, Personality, and Emotions in Social Media, (PEOPLES@COLING 2020)}.
  Association for Computational Linguistics.

\bibitem[\protect\citename{Biel and Gatica-Perez}2013]{biel2013youtube}
Joan-Isaac Biel and Daniel Gatica-Perez.
\newblock 2013.
\newblock The youtube lens: Crowdsourced personality impressions and
  audiovisual analysis of vlogs.
\newblock {\em Multimedia, IEEE Transactions on}, 15(1):41--55.

\bibitem[\protect\citename{Celli \bgroup et al.\egroup
  }2013]{celli2013workshop}
Fabio Celli, Fabio Pianesi, David Stillwell, and Michal Kosinski.
\newblock 2013.
\newblock Workshop on computational personality recognition: Shared task.
\newblock In {\em Seventh International AAAI Conference on Weblogs and Social
  Media}.

\bibitem[\protect\citename{Emmery \bgroup et al.\egroup
  }2017]{emmery-etal-2017-simple}
Chris Emmery, Grzegorz Chrupa{\l}a, and Walter Daelemans.
\newblock 2017.
\newblock Simple queries as distant labels for predicting gender on {T}witter.
\newblock In {\em Proceedings of the 3rd Workshop on Noisy User-generated
  Text}, pages 50--55, Copenhagen, Denmark, September. Association for
  Computational Linguistics.

\bibitem[\protect\citename{Go \bgroup et al.\egroup }2009]{go2009twitter}
Alec Go, Richa Bhayani, and Lei Huang.
\newblock 2009.
\newblock Twitter sentiment classification using distant supervision.
\newblock {\em CS224N project report, Stanford}, 1(12):2009.

\bibitem[\protect\citename{Goby}2006]{PersonalityandonlineofflinechoicesMbtiprofiles}
Valerie Goby.
\newblock 2006.
\newblock {Personality and Online/Offline Choices: MBTI Profiles and Favored
  Communication Modes in a Singapore Study}.
\newblock {\em Cyberpsychology \& behavior : the impact of the Internet,
  multimedia and virtual reality on behavior and society}, 9:5--13, 03.

\bibitem[\protect\citename{John and Srivastava}1999]{John1999TheBF}
Oliver~P. John and Sanjay Srivastava.
\newblock 1999.
\newblock The big five trait taxonomy: History, measurement, and theoretical
  perspectives.
\newblock In L.~A. Pervin and O.~P. John, editors, {\em Handbook of
  personality: Theory and research}, page 102–138. Guilford Press.

\bibitem[\protect\citename{Litvinova \bgroup et al.\egroup }2016]{litvinova}
Tatiana Litvinova, P.~Seredin, Olga Litvinova, and Olga Zagorovskaya.
\newblock 2016.
\newblock Profiling a set of personality traits of text author: What our words
  reveal about us.
\newblock {\em Research in Language}, 14, 12.

\bibitem[\protect\citename{Mairesse \bgroup et al.\egroup }2007]{mairasse}
Fran{\c c}ois Mairesse, {Marilyn A.} Walker, {Matthias R.} Mehl, and {Roger K.}
  Moore.
\newblock 2007.
\newblock Using linguistic cues for the automatic recognition of personality in
  conversation and text.
\newblock {\em Journal of Artificial Intelligence Research}, 30:457--500, sep.

\bibitem[\protect\citename{Mehta \bgroup et al.\egroup }2019]{mehta2019recent}
Yash Mehta, Navonil Majumder, Alexander Gelbukh, and Erik Cambria.
\newblock 2019.
\newblock Recent trends in deep learning based personality detection.
\newblock {\em Artificial Intelligence Review}, pages 1--27.

\bibitem[\protect\citename{Myers and Myers}1995]{myers1995gifts}
I.B. Myers and P.B. Myers.
\newblock 1995.
\newblock {\em Gifts Differing: Understanding Personality Type}.
\newblock Mobius.

\bibitem[\protect\citename{Nieuwenhuis and Nissim}2019]{embYT}
Moniek Nieuwenhuis and Malvina Nissim.
\newblock 2019.
\newblock {The Contribution of Embeddings to Sentiment Analysis on YouTube}.
\newblock In {\em Proceedings of the Sixth Italian Conference on Computational
  Linguistics, Bari, Italy, November 13-15, 2019}, volume 2481 of {\em {CEUR}
  Workshop Proceedings}. CEUR-WS.org.

\bibitem[\protect\citename{Pardo \bgroup et al.\egroup
  }2015]{Pardo2015OverviewOT}
Francisco M.~Rangel Pardo, Fabio Celli, Paolo Rosso, Martin Potthast, Benno
  Stein, and Walter Daelemans.
\newblock 2015.
\newblock {Overview of the 3rd Author Profiling Task at PAN 2015}.
\newblock In {\em Working Notes of {CLEF} 2015 - Conference and Labs of the
  Evaluation forum, Toulouse, France, September 8-11, 2015}, volume 1391 of
  {\em {CEUR} Workshop Proceedings}. CEUR-WS.org.

\bibitem[\protect\citename{Parks and Guay}2009]{parks2009personality}
Laura Parks and Russell~P Guay.
\newblock 2009.
\newblock Personality, values, and motivation.
\newblock {\em Personality and individual differences}, 47(7):675--684.

\bibitem[\protect\citename{Pedregosa \bgroup et al.\egroup }2011]{scikit-learn}
F.~Pedregosa, G.~Varoquaux, A.~Gramfort, V.~Michel, B.~Thirion, O.~Grisel,
  M.~Blondel, P.~Prettenhofer, R.~Weiss, V.~Dubourg, J.~Vanderplas, A.~Passos,
  D.~Cournapeau, M.~Brucher, M.~Perrot, and E.~Duchesnay.
\newblock 2011.
\newblock Scikit-learn: Machine learning in {P}ython.
\newblock {\em Journal of Machine Learning Research}, 12:2825--2830.

\bibitem[\protect\citename{Pennebaker and King}2000]{essays}
James Pennebaker and Laura King.
\newblock 2000.
\newblock Linguistic styles: Language use as an individual difference.
\newblock {\em Journal of personality and social psychology}, 77:1296--312, 01.

\bibitem[\protect\citename{Pennington \bgroup et al.\egroup }2014]{embGloVe}
Jeffrey Pennington, Richard Socher, and Christopher~D. Manning.
\newblock 2014.
\newblock Glove: Global vectors for word representation.
\newblock In {\em Empirical Methods in Natural Language Processing (EMNLP)},
  pages 1532--1543.

\bibitem[\protect\citename{Plank and Hovy}2015]{Personality_Traits_on_Twitter}
Barbara Plank and Dirk Hovy.
\newblock 2015.
\newblock Personality traits on {T}witter{---}or{---}{H}ow to get 1,500
  personality tests in a week.
\newblock In {\em Proceedings of the 6th Workshop on Computational Approaches
  to Subjectivity, Sentiment and Social Media Analysis}, pages 92--98, Lisboa,
  Portugal, September. Association for Computational Linguistics.

\bibitem[\protect\citename{Pool and Nissim}2016]{pool2016distant}
Chris Pool and Malvina Nissim.
\newblock 2016.
\newblock Distant supervision for emotion detection using {F}acebook reactions.
\newblock In {\em Proceedings of the Workshop on Computational Modeling of
  People{'}s Opinions, Personality, and Emotions in Social Media ({PEOPLES})},
  pages 30--39, Osaka, Japan, December. The COLING 2016 Organizing Committee.

\bibitem[\protect\citename{Schwartz \bgroup et al.\egroup
  }2013]{schwartz2013personality}
H.~Andrew Schwartz, Johannes~C. Eichstaedt, Margaret~L. Kern, Lukasz
  Dziurzynski, Stephanie~M. Ramones, Megha Agrawal, Achal Shah, Michal
  Kosinski, David Stillwell, Martin~E.P. Seligman, et~al.
\newblock 2013.
\newblock Personality, gender, and age in the language of social media: The
  open-vocabulary approach.
\newblock {\em PloS one}, 8(9):e73791.

\bibitem[\protect\citename{Snyder}1983]{snyder1983influence}
Mark Snyder.
\newblock 1983.
\newblock The influence of individuals on situations: Implications for
  understanding the links between personality and social behavior.
\newblock {\em Journal of personality}, 51(3):497--516.

\bibitem[\protect\citename{Staiano \bgroup et al.\egroup }2012]{mobileDataset}
Jacopo Staiano, Bruno Lepri, Nadav Aharony, Fabio Pianesi, Nicu Sebe, and Alex
  Pentland.
\newblock 2012.
\newblock Friends don't lie - inferring personality traits from social network
  structure.
\newblock In {\em UbiComp'12 - Proceedings of the 2012 ACM Conference on
  Ubiquitous Computing}, pages 321--330, 09.

\bibitem[\protect\citename{Verhoeven \bgroup et al.\egroup
  }2016]{Verhoeven2016TwiStyAM}
Ben Verhoeven, Walter Daelemans, and Barbara Plank.
\newblock 2016.
\newblock {T}wi{S}ty: A multilingual {T}witter stylometry corpus for gender and
  personality profiling.
\newblock In {\em Proceedings of the Tenth International Conference on Language
  Resources and Evaluation ({LREC}'16)}, pages 1632--1637, Portoro{\v{z}},
  Slovenia, May. European Language Resources Association (ELRA).

\bibitem[\protect\citename{Vinciarelli and
  Mohammadi}2014]{vinciarelli2014survey}
Alessandro Vinciarelli and Gelareh Mohammadi.
\newblock 2014.
\newblock A survey of personality computing.
\newblock {\em IEEE Transactions on Affective Computing}, 5(3):273--291.

\bibitem[\protect\citename{Weiner and Greene}2017]{ethicalconsiderations}
Irving~B. Weiner and Roger~L. Greene, 2017.
\newblock {\em Ethical Considerations In Personality Assessment}, chapter~4,
  pages 59--74.
\newblock Wiley.

\bibitem[\protect\citename{Whelan and Davies}2006]{whelan2006profiling}
Susan Whelan and Gary Davies.
\newblock 2006.
\newblock Profiling consumers of own brands and national brands using human
  personality.
\newblock {\em Journal of Retailing and Consumer Services}, 13(6):393--402.

\bibitem[\protect\citename{Youyou \bgroup et al.\egroup
  }2015]{youyou2015computer}
Wu~Youyou, Michal Kosinski, and David Stillwell.
\newblock 2015.
\newblock Computer-based personality judgments are more accurate than those
  made by humans.
\newblock {\em Proceedings of the National Academy of Sciences},
  112(4):1036--1040.

\end{thebibliography}

\end{document}